%% file: lipz-sys-nips.tex
\documentclass{article}

\usepackage[nonatbib,final]{nips_2018}
\usepackage{matharticle}
\usepackage{paralist}
\usepackage{todonotes}
\usepackage{xspace}
\usepackage{subfig}

\newcommand{\horse}{\texttt{Lipizzaner}\xspace}

\newcommand{\framework}{system\xspace}   %system
\newcommand{\Framework}{System\xspace}

\usepackage[utf8]{inputenc} % allow utf-8 input
\usepackage[T1]{fontenc}    % use 8-bit T1 fonts
\usepackage{hyperref}       % hyperlinks
\usepackage{url}            % simple URL typesetting
\usepackage{booktabs}       % professional-quality tables
\usepackage{amsfonts}       % blackboard math symbols
\usepackage{nicefrac}       % compact symbols for 1/2, etc.
\usepackage{microtype}      % microtypography
\usepackage{comment}
\usepackage[numbers]{natbib}
\usepackage{wrapfig}

\algblock{ParFor}{EndParFor}
\algnewcommand\algorithmicparfor{\textbf{parfor}}
\algnewcommand\algorithmicpardo{\textbf{do}}
\algnewcommand\algorithmicendparfor{\textbf{end\ parfor}}
\algrenewtext{ParFor}[1]{\algorithmicparfor\ #1\ \algorithmicpardo}
\algrenewtext{EndParFor}{\algorithmicendparfor}

\title{Lipizzaner: A System That Scales Robust Generative Adversarial Network Training}

\author{
Tom Schmiedlechner\\
  CSAIL, MIT, USA\\
\texttt{schmied@mit.edu}
\And 
Ignavier Ng Zhi Yong\\
  CSAIL, MIT, USA\\
  \texttt{ignavier@mit.edu}
\And Abdullah Al-Dujaili\\
  CSAIL, MIT, USA\\
  \texttt{aldujail@mit.edu}
\And Erik Hemberg\\
  CSAIL, MIT, USA\\
  \texttt{hembergerik@csail.mit.edu}
\And  Una-May O’Reilly \\
  CSAIL, MIT, USA\\
 \texttt{unamay@csail.mit.edu} \\
}

\begin{document}

\maketitle

\input{abstract}

\input{introduction}
\input{background}

\input{methods}
\input{experiments}

\input{conclusion}

\begin{comment}
\subsubsection*{Acknowledgments}

Use unnumbered third level headings for the acknowledgments. All
acknowledgments go at the end of the paper. Do not include
acknowledgments in the anonymized submission, only in the final paper.

\end{comment}

\bibliographystyle{plainnat}
{
	\small
	\bibliography{bibliography}
}
\end{document}

%% file: abstract.tex
\begin{abstract}
  GANs are difficult to train due to convergence pathologies such as mode and
  discriminator collapse. We introduce \horse, an open source software \framework that allows machine learning engineers to train GANs in a distributed and robust way. \horse distributes a competitive coevolutionary algorithm which, by virtue of dual, adapting, generator and discriminator populations, is robust to collapses. The algorithm is well suited to efficient distribution because it uses a spatial grid abstraction. Training is local to each cell and strong intermediate training results are exchanged among overlapping neighborhoods allowing high performing solutions to propagate and improve with more rounds of training.  Experiments on common image datasets overcome  critical collapses. Communication overhead scales linearly when increasing the number of compute instances and we observe that increasing scale leads to improved model performance.
\end{abstract}

%% file: introduction.tex
\section{Introduction}
\label{sec:intro}
%TODO why, what, how (Missing why)

	Despite their demonstrated success, it is well known that Generative Adversarial Networks (GANs) are difficult to train. The objective of training is to derive a generator that is able to completely  thwart  
	the discriminator in its ability to identify genuine samples from ones offered by the generator. GAN training can be formulated as a two-player minimax game: the (neural network) discriminator is trying to maximize its payoff (accuracy), and the (neural network) generator  is trying to minimize the discriminator's payoff (accuracy).  
	The two networks are differentiable, and therefore optimizing them is achieved by simultaneous gradient-based updates to their parameters. In practice, gradient-based GAN training often converges to payoffs that are sub-optimally stuck in oscillation or collapse. This is partly  because gradient-based updates seek a  stationary solution with zero gradient. 	This 
objective is a necessary condition for a single network to converge, but in the case of the GAN's coupled optimization, equilibrium is the corresponding necessary condition for convergence.  Consequently, a variety of degenerate training behaviors has been observed---e.g., \emph{mode collapse}~\cite{Arora2017GANs}, \emph{discriminator collapse}~\cite{Li2017Towards},  and \emph{vanishing gradients}~\cite{Arjovsky2017Towards}. 
These unstable learning dynamics have been a key limiting factor in training GANs in a robust way, let alone tuning their hyperparameters or scaling training. 
\citet{schmiedlechner2018towards} offer a robust training solution that combines the training of multiple GANs with grid-based competitive coevolution.  Succinctly, the training of each GAN pair is done with stochastic gradient descent (SGD) while the grid-based coevolutionary algorithm adaptively selects higher performing models for iterative training by referencing the GANs in cells in a local neighborhood. Overlapping neighborhoods and local communication allow efficient propagation of improving models. The impressive performance of this solution prompts us to distribute it, see Figure~\ref{fig:distributed-arch}~(a), so that training is faster (by wall clock timing) and efficiently scalable.  Our contribution is a scalable, parallelized and distributed GAN training \framework, implemented and licensed as open source\footnote{{\url{https://github.com/ALFA-group/lipizzaner-gan}}}, based on \cite{schmiedlechner2018towards}'s solution. Additionally, the scaling allows us to experimentally determine that larger grids (i.e. more spatially distributed GAN training) yield better trained GANs.

%We proceed as follows. As background and related work, we discuss GAN train challenges, coevolutionary algorithms and their scaling. We then describe the~\horse \framework and experimentally demonstrate its performance and scaling properties. We conclude with future work.

%% file: background.tex
\section{Background}
\label{sec:related}

\textit{Improving GAN Training.} Robust GAN training is still an open
research topic~\cite{Arora2017Generalization}. Simple theoretical GAN models have been proposed to provide a better understanding of the problem~\cite{Li2017Towards}. For algorithmic implementations, several tips and tricks have been suggested to stabilize the training over the past years~\cite{ganhacks}. Some use
hard-coded conditions to decrease the optimizers' learning rate after
a given number of iterations~\cite{Radford2015unsupervised}, while others employ
ensemble concepts~\cite{wang2016ensembles}. Motivated by the similarity of degenerate behaviors in GAN training to the decade-old observed patterns in competitive coevolutionary algorithm dynamics (i.e., loss of gradient, focusing, and relativism), \citet{schmiedlechner2018towards} propose a spatial coevolution approach for GAN training. The authors conduct experiments on the theoretical GAN model of~\cite{Li2017Towards}. They show, using the theoretical model, that a basic coevolutionary algorithm with Gaussian-based mutations can escape behaviors such as mode and discriminator collapse. They also run a small-scale spatial coevolution with gradient-based mutations to update the neural net parameters and Gaussian-based mutations to update the hyperparameters on the MNIST and CelebA datasets.  The bulk of attempts for improving GAN training have been designed to fit a single machine (or a single GPU). The advent of large-scale parallel computation infrastructure prompts our interest in scaling them.  To do so, we select \cite{schmiedlechner2018towards}'s solution because of its use of evolutionary computing. 

\textit{Evolutionary Computing.}  Evolutionary algorithms are
population-based optimization techniques. %They evolve a population of
%{candidate solutions} using different operators to create a new generation
%from the best-rated solutions of the previous generation. To do so,
%each solution is evaluated against a fitness function. Then, a
%number of parent individuals are selected with a fitness bias to populate the next
%generation~\cite{Back1996EA}.
 Competitive coevolutionary algorithms
have adversarial populations (usually two) that simultaneously
evolve~\cite{Hillis1990Coev} population solutions against each other. Unlike classic evolutionary algorithms,
they employ fitness functions that rate solutions relative to their
\emph{opponent} population.  Formally, these algorithms can be
described with a minimax formulation~\cite{Herrmann1999Genetic,Ash2018Reckless}, and therefore share common
characteristics with GANs. 

\textit{Scaling Evolutionary Computing for ML.}  A team from \textsf{OpenAI}~\cite{salimans2017evolution} applied a simplified version of Natural Evolution Strategies~(NES)~\cite{wierstra2008NES} with a novel communication strategy to a collection of reinforcement learning (RL) benchmark problems. Due to better parallelization over thousand cores, they achieved much faster training times (wall-clock time) than popular RL techniques. Likewise, a team from~\textsf{Uber~AI}~\cite{clune2017uber}
%\cite{such2017deep, conti2017improving}
showed that deep convolutional networks with over $4$ million parameters trained with genetic algorithms can also reach results competitive to those trained with \textsf{OpenAI}'s NES and other RL algorithms. \textsf{OpenAI} ran their experiments on a computing cluster of $80$ machines and $1440$ CPU cores~\cite{salimans2017evolution}, whereas \textsf{Uber~AI} employed a range of hundreds to thousands of CPU cores (depending on availability). Another effective mean to scale up evolutionary algorithm in a distributed setting is spatial (toroidal) coevolution, which controls the mixing of adversarial populations in coevolutionary algorithms.  The
members of populations are divided up on a grid of cells and each cell has a local neighborhood. A neighborhood is defined by adjacent cells and  specified by its size, $n_{cells}$. This reduces the cost of overall
communication from $\mathcal{O}(n^2)$ to $\mathcal{O}(n_{cells}n)$,
where $n$ is the size of each population. Five cells
per neighborhood (one center and four adjacent cells) are
common~\cite{Husbands1994Distributed}. With this notion of distributed
evolution, each neighborhood can evolve in a different direction and
more diverse points in the search
space are explored~\cite{Mitchell2006Coevolutionary,Williams2005Investigating}.  The next section presents \horse: a scalable, distributed \framework for coevolutionary GAN training.

%% file: methods.tex
\section{The \horse \Framework}
\label{sec:methods}

We start with a brief description of how \horse trains. Second, we discuss the general design principles and requirements for a scalable architecture of the framework. Then, we describe the concrete implementation steps of the resulting system.

\begin{figure}
\begin{tabular}{cc}
	\raisebox{+1.2\height}{
	\includegraphics[width=0.28\textwidth]{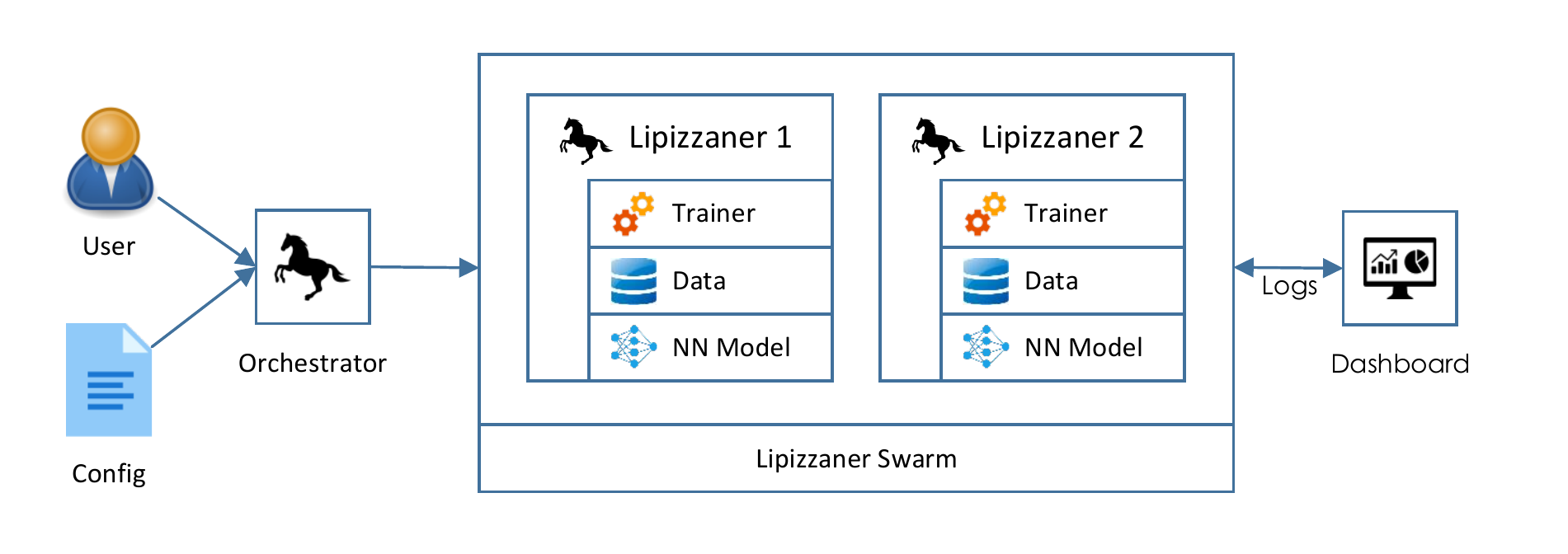}
}
	& 
	\includegraphics[width=0.68\textwidth]{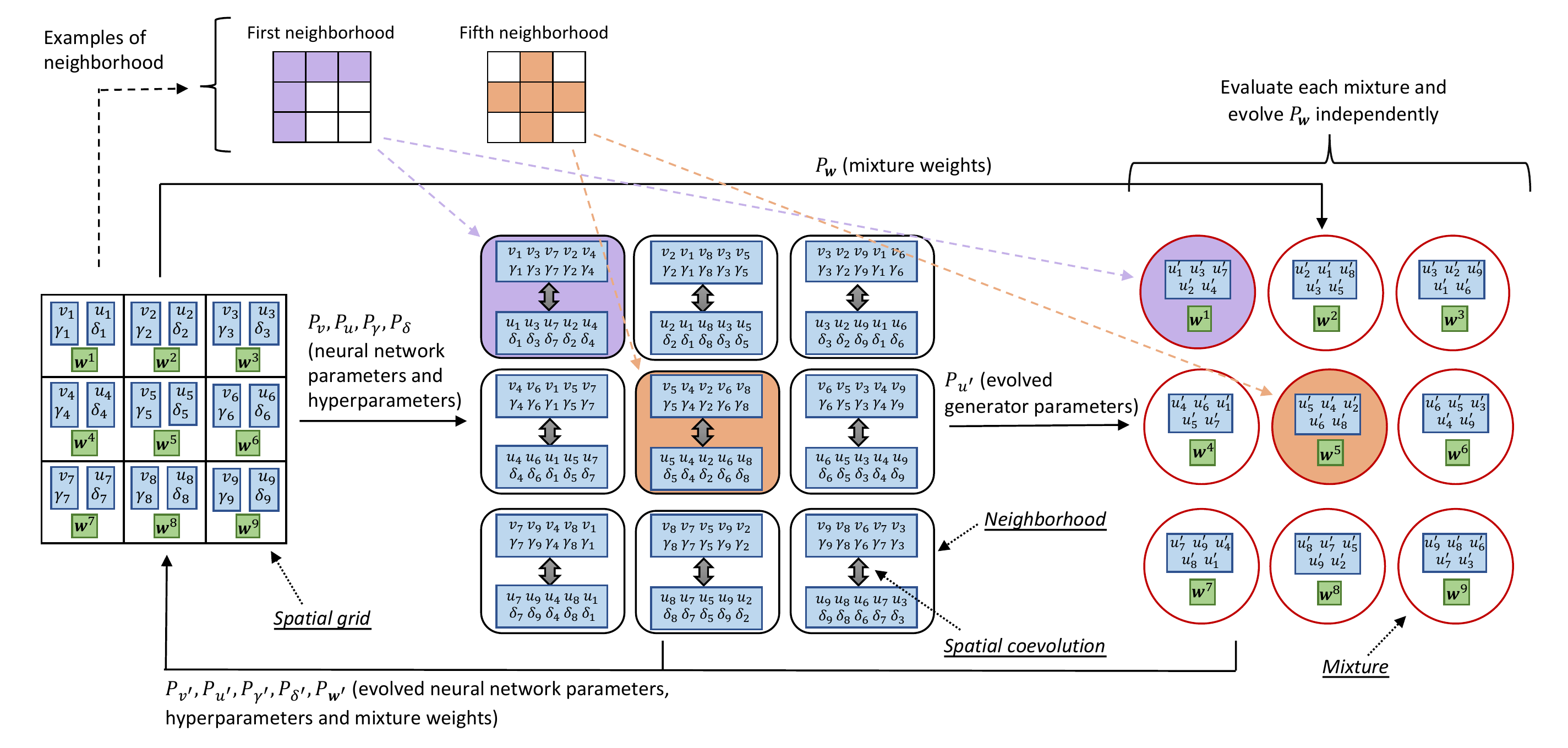}
	\\
	(a) & (b)
\end{tabular}
\caption{(a) \footnotesize High-level view of \horse's architecture. User provides
	configuration file to the \textit{orchestrator}. A distributed
	\horse swarm is controlled by the \textit{orchestrator}. Each node
	in the swarm asynchronously trains the combination of the cell's GANs with its neighbors'. 
	The dashboard shows progress and
	results. (b) \footnotesize \horse training on a
	$3 \times 3$ grid. $P_v = \{v_1, \dots v_9\}$ and $P_v = \{u_1, \dots
	u_9\}$ denote neural network parameters of discriminator and
	generator population respectively. $P_{\gamma} = \{\gamma_1,
	\dots, \gamma_9\}$ and $P_{\delta} = \{\delta_1, \dots, \delta_9\}$
	denote the hyperparameters (e.g., learning rate) of discriminator
	and generator population, respectively. $P_{\bf w} = \{\mathbf{w}_1, \dots, \mathbf{w}_9\}$
	denote the mixture weights. The $(\cdot)'$ notation denotes the  value of $(\cdot)$ after one iteration of (co)evolution.}
	 \label{fig:distributed-arch}
\end{figure}

\textit{Coevolutionary GAN Training.} The coevolutionary framework is executed in an asynchronous fashion as described in the following steps (depicted in Figure~\ref{fig:distributed-arch}~(b)). 
\begin{inparaenum}[\itshape 1)]
	\item Randomly initialize each cell in the grid with a  generator
	and a discriminator of random weight and hyperparameters. The $i$-th generator is described by its neural net parameters $u_i$ and hyperparameters $\delta_i$. The $j$-th discriminator is similarly described with $v_j$ and $\gamma_j$. The generators (discriminators) from all the cells form the generator (discriminator) population $P_u, P_\delta$ ($P_v, P_\gamma$).
	\item Each generator (discriminator) is evaluated against each of the discriminators (generators) in its neighborhood. The evaluation process computes $\mathcal{L}(u_i, v_j)$: the value of the GAN objective (loss function) $\mathcal{L}$ at the corresponding generator $u_i$ and discriminator $v_j$. The values of a discriminator's (generator's) interactions are averaged (negative-averaged) to constitute its fitness.
	\item Generators and discriminators in each neighborhood are selected based on
    tournament selection. For the selected generator and discriminator, SGD training then performs gradient-based updates on their neural net parameters $u_i$ and $v_j$, while Gaussian-based updates create new hyperparameters values $\delta_i'$ and $\gamma_j'$. 
	\item \horse produces a mixture of generators. It assigns a mixture weight vector $\mathbf{w}$ for each neighborhood. These vectors are evolved using an \texttt{ES}-(1+1) algorithm which optimizes for the performance (e.g., Fr\'echet Inception Distance (FID) score~\cite{Martin2017GANs}) of the neighborhood generators (weighted by $\mathbf{w}$).
	\item Go to step \textit{2)}.
\end{inparaenum}
We next describe \horse's two system level modules.

\begin{comment}
\begin{figure}
\begin{minipage}[t]{0.3\textwidth}
	\includegraphics[width=\textwidth]{figures/seabiscuit-distributed-architecture}
	\captionof{figure}{\footnotesize High-level view of \horse's architecture. User provides
		configuration file to the \textit{orchestrator}. A distributed
		\horse swarm is controlled by the \textit{orchestrator}. Each node
		in the swarm asynchronously trains the combination of the cell's GANs with its neighbors'. 
		The dashboard shows progress and
		results.}
	\label{fig:distributed-arch}
\end{minipage}\hspace{0.5em}
\begin{minipage}[t]{0.7\textwidth}
	\centering
	\includegraphics[width=\textwidth]{figures/lipz_alg_schema2.pdf}
	\caption{\small \horse training on a
		$3 \times 3$ toroidal grid. $P_v = \{v_1, \dots v_9\}$ and $P_v = \{u_1, \dots
		u_9\}$ denote neural network parameters of discriminator and
		generator population respectively. $P_{\gamma} = \{\gamma_1,
		\dots, \gamma_9\}$ and $P_{\delta} = \{\delta_1, \dots, \delta_9\}$
		denote the hyperparameters (e.g., learning rate) values of discriminator
		and generator population, respectively. $P_{\bf w} = \{\mathbf{w}_1, \dots, \mathbf{w}_9\}$
		denote the mixture weights. The $(\cdot)'$ notation denotes the  value of $(\cdot)$ after one iteration of (co)evolution.}
	\label{fig:lipz_alg_schema}
\end{minipage}
\end{figure}
\end{comment}

\textit{System Modules.}
As shown in Figure~\ref{fig:distributed-arch}~(a), \horse has a core and a dashboard module. 
\begin{inparaenum}[\itshape 1)]
\item The core module has a component-based design. All components exist on
each distributed \horse instance. They are:
\begin{inparaitem}
\item [\textit{Input data loader}] that manages the data samples for
  training, i.e. the distribution the generator tries to reproduce.
\item [\textit{Neural network model}] that generates or discriminates data.
\item [\textit{Trainer}] that executes the training iterations\footnote{Iteration, generation and epoch are used interchangeably in our system} of the
  evolutionary process itself. \textit{Trainer} accesses input data and the models from their respective components and evolves them  with the settings provided by the configuration.
\item [\textit{Distribution server and client}] that sends and receives
  data via a TCP/IP interface. The server component offers a public
  API and endpoints for accessing the state of an instance. The
  fitness values of the individuals and the internal state of the
  gradient optimizers are shared. The state of some optimizers
  is lightweight, while e.g. Adam requires transmission of complex
  state objects.
\item [\textit{Configuration}] that is vertically aligned over the system
  and connected to all components. All parameters are specified in
  configuration files. This reduces redeployment and code editing.
\end{inparaitem}
\item \horse's dashboard module is designed to simplify the analysis of
experiments and make training transparent. The GUI-based monitoring shows phenomena like solutions
propagating through the grid, oscillation between generators and
discriminators, etc. GUI
components and their interactions are illustrated in
Figure~\ref{fig:dashboard}.  The \textit{Log database} contains
details about the executed experiments and instances.  The
\textit{Back-end controller} is a server-side component connecting the
front-end to the log database. Finally, \textit{Front-end component
and view} contains the logic to access the experiment and result
data from the back-end controller and to inject it into the view.
\end{inparaenum}

\textit{Implementation and Distribution.} The core module is
written in \texttt{Python3} and uses
\texttt{pytorch}\footnote{\url{https://pytorch.org/}}. The dashboard
is an \texttt{Angular.js} single page web application using
\texttt{Type Script} and \texttt{ASP.NET Core}. The trainer component defines and hosts the executed evolutionary
process. Currently \horse supports different types of trainers;
gradient-free trainers for evolution strategies and Natural Evolution
Strategies~\cite{wierstra2008NES}, and trainers that update the
neural net parameters with gradient-based optimizers.
Most GAN types primarily differ by the way they update their weights
and fitness evaluation. \horse injects the necessary
functionality into the respective trainer (as most training
functionality is not class-specific, but based on interfaces passed to
the constructor, i.e. \emph{Dependency Injection}). 
The precise training steps differ slightly depending on the trainer and type of
GAN, but all share a common coevolutionary baseline
procedure.

\begin{minipage}{0.5\textwidth}
 To support distribution, \horse uses
\texttt{Docker}\footnotemark swarm and
\texttt{docker-machine}. A master-client design pattern
is used for communication. Clients are added by starting the
application on more nodes, running pre-configured virtual machines, or
with \texttt{Docker}. 

The \horse master (or \emph{orchestrator}) is meant to
  control a single experiment. Its tasks are:
\begin{inparaenum}[\itshape 1)]
\item parse the configuration and connect to clients and transmits the
  experiment to all of them.
\item periodically check client state, i.e. if the experiment has
ended, or if the client may be not reachable anymore, unreachable
  clients can be ignored, or the entire experiment terminated.
\item gather finished results, save them to its disk and rank the
  final mixtures by their scores. Create sample images. \end{inparaenum}
The task overhead requires only modest computation power. \horse instances communicate with HTTP web services and exchange only
relatively small amounts of data during the training process, it is
possible to deploy multiple instances onto different machines and
hence scale horizontally.

From a logical perspective, each client represents one cell in the
spatial grid topology. An experiment request contains all
configuration options a user pass into the master 
application, as it
is forwarded to the client.
The typical behavior of the client has three steps:
\begin{inparaenum}[\itshape 1)]
	\item If no experiment is running, the client runs in the background
	and listens for experiment requests on a specific port. When an
	experiment is requested, the client parses the received
	configuration file and executes the specified training algorithm. It
	furthermore requests data from the neighboring cells each time the
	algorithm accesses the respective properties of the populations.
	\item It offers HTTP endpoints simultaneously to execute the training
	process. Other clients can also access these endpoints and request
	the current populations and optimizer parameters.
	\item The master actively monitors the clients and collects the
	results. After this, the client changes its state from \texttt{Busy}
	to \texttt{Idle} and waits for new experiment requests. 
\end{inparaenum}
\end{minipage}\hspace{0.5em}
\begin{minipage}{0.46\textwidth}
	\centering
	\includegraphics[width=\textwidth]{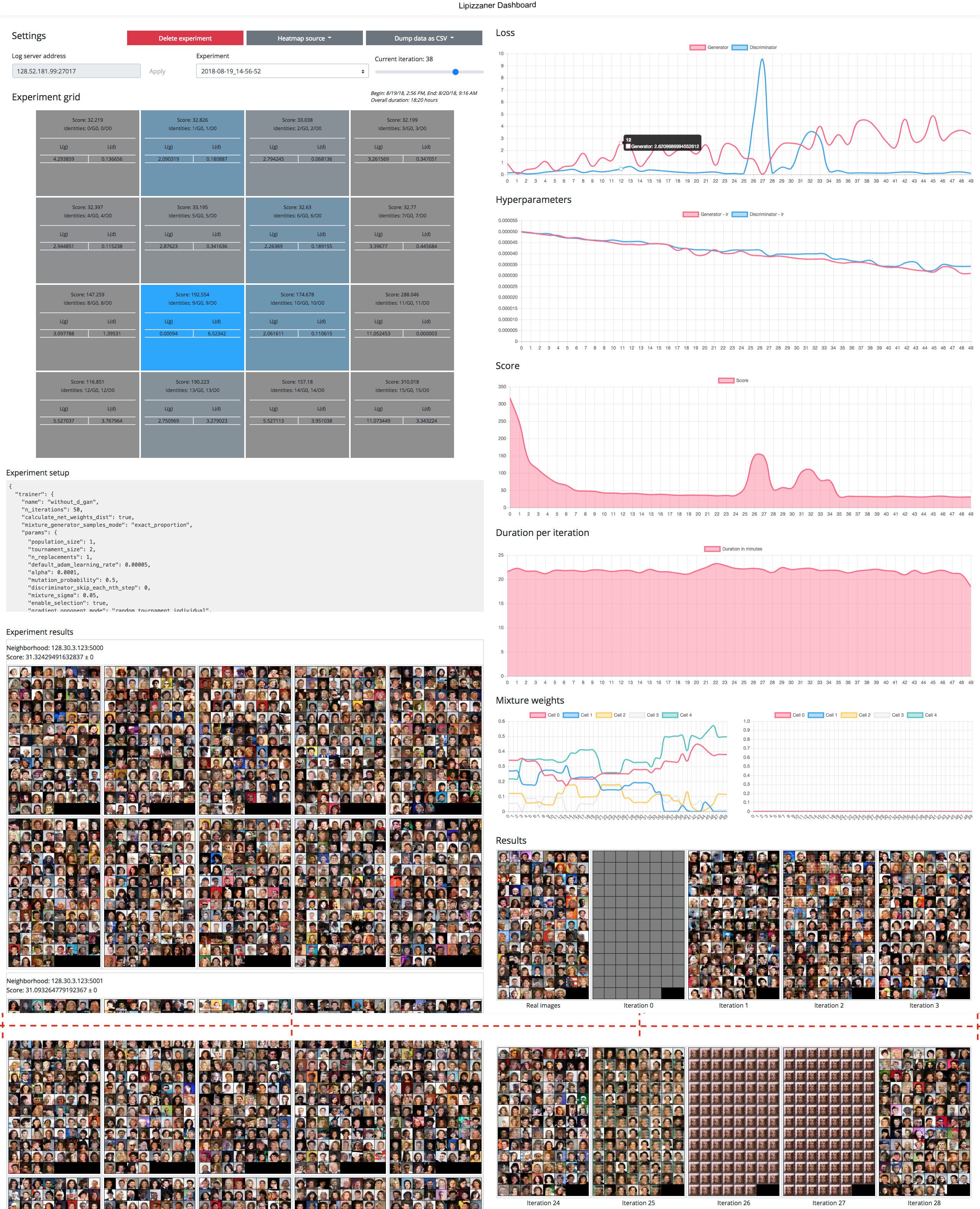}
	\captionof{figure}{\footnotesize{Screen shot of the \horse dashboard web
			application---a demo dashboard page can be found at \url{https://github.com/ALFA-group/lipizzaner-gan/blob/master/dashboard-demo/dashboard.html} for a better readibility. The orange dashed line
			indicates that images from iteration 4 to 23 are cropped.  The
			navigation component is on the left, the details component is on
			the right side of the screen. The {navigation} loads and
			displays the experiment selection dialog elements. For a
			selected experiment, configuration, details, topology, and
			execution time are shown. For an experiment done, samples from
			the resulting mixtures are shown as well. It is possible to
			scroll through training iterations while displaying a live heat
			map of the grid. When a grid cell of a specific experiment is
			selected, the {details} component to the right displays
			drill-down information about the whole experiment history of the
			cell. This includes charts for loss, hyperparameters, mixture
			weights and score values. Intermediate generator output images
			for each iterations are displayed as well, together with real
			images from the input dataset.}}
	\label{fig:dashboard}
	% Abit of hardcoding for the footnote, refer to https://tex.stackexchange.com/a/10185
	% Must double check with the superscript digit every time when modifying footnote
\end{minipage}

\footnotetext{\url{https://www.docker.com}}

%% file: experiments.tex
\section{Experiments}
\label{sec:experiments}
This section provides empirical evaluation of \horse on two common image datasets, MNIST and CelebA. We assess the system in terms of its scaling properties and generative modeling performance. The settings used for the experiments with
\horse are shown in Table~\ref{tbl:exp-lpz}.

\subsection{MNIST Dataset}

\begin{figure}
	\centering
	\begin{minipage}[t][5cm]{.45\textwidth}
		\centering
		\includegraphics[width=1\linewidth]{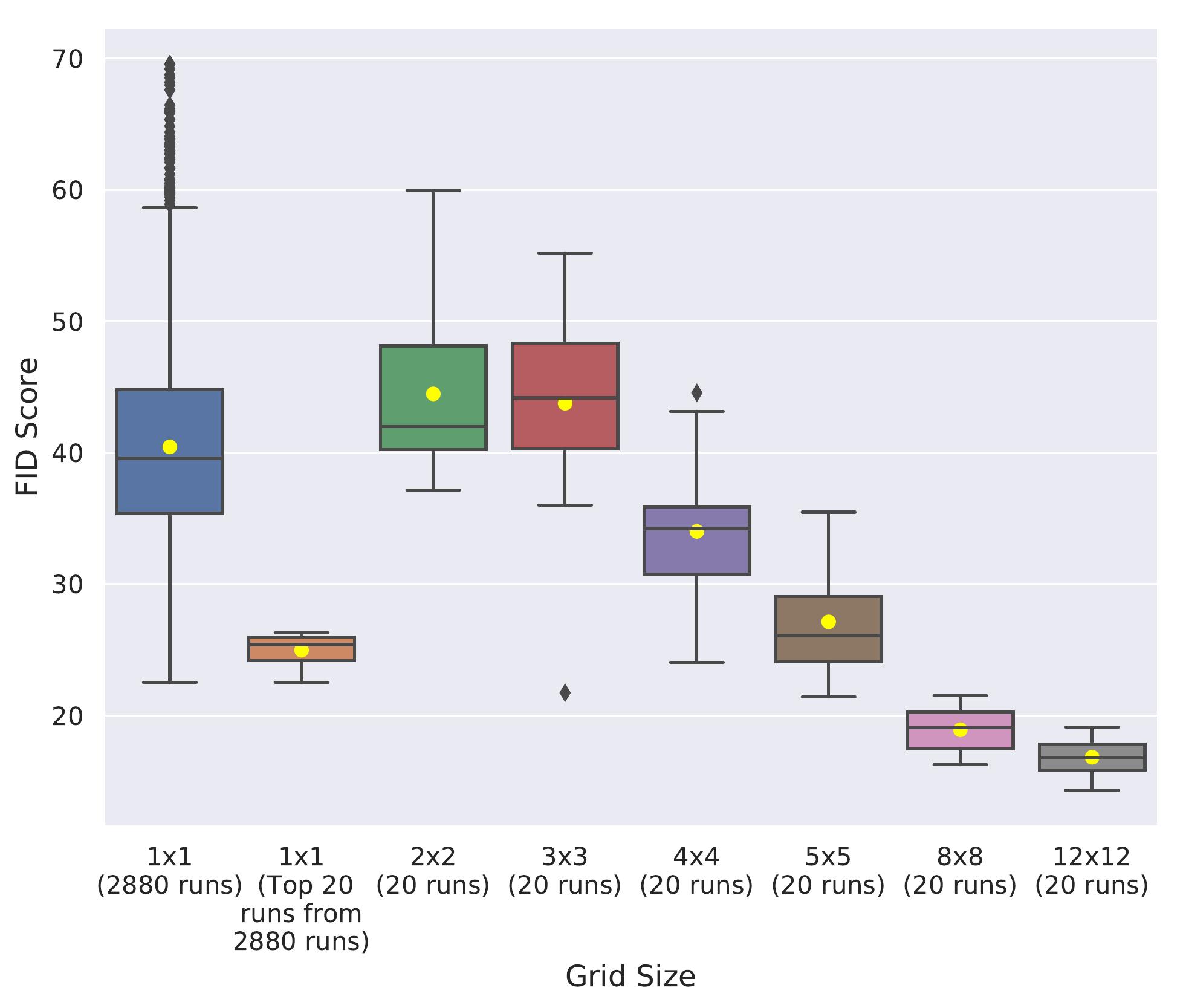}
		\captionof{figure}{\footnotesize{Box plot of the FID score for different grid sizes on
				MNIST. The x-axis shows box plots for different grid sizes and the
				y-axis shows the FID score. Yellow dots are the average FID, black
				diamonds are outliers. Larger grid sizes have lower FID
				scores. For the sake of legibility, only those experiments having
				FID score lower than 70 are included in the box plot. For $1 \times 1$
				grid, there are 27, 7, 18 outliers which lie in the interval  of (70, 100),
				(100, 140), (500, 1100) respectively.}}
		\label{fig:MNIST_FID}
	\end{minipage}%
	\hspace{1em}
	\begin{minipage}[b][2cm]{.45\textwidth}
		\centering
		\captionof{table}{\footnotesize Setup for experiments conducted with the \horse
			system on MNIST and CelebA datasets.}
		\label{tbl:exp-lpz}
		\resizebox{\textwidth}{!}{
			\begin{tabular}{l|l|l}
				\toprule
				\textbf{Parameter} & \textbf{MNIST} & \textbf{CelebA} \\
				\toprule
				\multicolumn{3}{c}{\textbf{Coevolutionary settings}} \\ \midrule
				Iterations & 200 & 50 \\ \midrule
				Population size per cell & 1 & 1 \\ \midrule
				Tournament size & 2 & 2\\ \hline
				Grid size & $1 \times 1$ to $12 \times 12$ & $1 \times 1$ to $4 \times 4$ \\ \hline
				Mixture mutation scale & 0.01 & 0.05 \\ \midrule
				\multicolumn{3}{c}{\textbf{Hyperparameter mutation}} \\ \midrule
				Optimizer & Adam & Adam \\ \midrule
				Initial learning rate & 0.0002 & 0.00005 \\ \midrule
				Mutation rate & 0.0001 & 0.0001 \\ \midrule
				Mutation probability & 0.5 & 0.5 \\ \midrule
				\multicolumn{3}{c}{\textbf{Network topology}} \\ \midrule
				Network type  & MLP & DCGAN \\ \midrule
				Input neurons & 64 & 100 \\ \midrule
				Number of hidden layers & 2 & 4 \\ \midrule
				Neurons per hidden layer & 256 & $16,384 - 131,072$ \\ \midrule
				Output neurons  & 784 & $64 \times 64 \times 3$ \\ \midrule
				Activation function & $\tanh$ & $\tanh$ \\ \midrule
				\multicolumn{3}{c}{\textbf{Training settings}} \\ \midrule
				Batch size  & 100 & 128 \\ \midrule
				Skip N disc. steps & 1 & - \\
				\bottomrule
			\end{tabular}
		}
	\end{minipage}
\end{figure}

\textit{Scalability and Performance.} \horse
improves the performance, convergence speed and stability of the
generator for larger grid sizes when measuring the average FID score
over multiple runs, see Figure~\ref{fig:MNIST_FID}. A rank sum test
with Bonferroni correction shows significant differences for the grid
sizes larger than $4 \times 4$ at 99\% confidence level. One
hypothesis for the behavior of \horse is that there is less overlap in
the neighborhoods for these grid sizes. We execute the $1 \times 1$ grid
for 2,880 runs in order to use similar compute effort as the $12 \times
12$ grid. Even then the minimum FID for the $1 \times 1$ grid (22.5)
is higher than the maximum FID of the $8 \times 8$ grid
(21.5). Multiple outliers and discriminator collapses are observed for
$1 \times 1$ grid, whereas larger grid sizes not only improve the
stability with smaller standard deviation and less outliers, but also
manage to completely overcome discriminator collapses. 
%The average FID score over all mixtures for each iteration is shown in Figure~\ref{fig:MNIST_convergence}. The larger grid sizes converge faster and have lower variance. 
These experiments were conducted on a
GPU cluster node which consists of eight Nvidia Titan Xp with 12 GB
RAM, 16 Intel Xeon cores with 2.2GHz each, and 125 GB RAM.

\begin{figure}[t]
	\begin{minipage}{\textwidth}
	\subfloat[\scriptsize{\tiny TVD for different grid sizes}]
	{\includegraphics[width=0.249\textwidth]{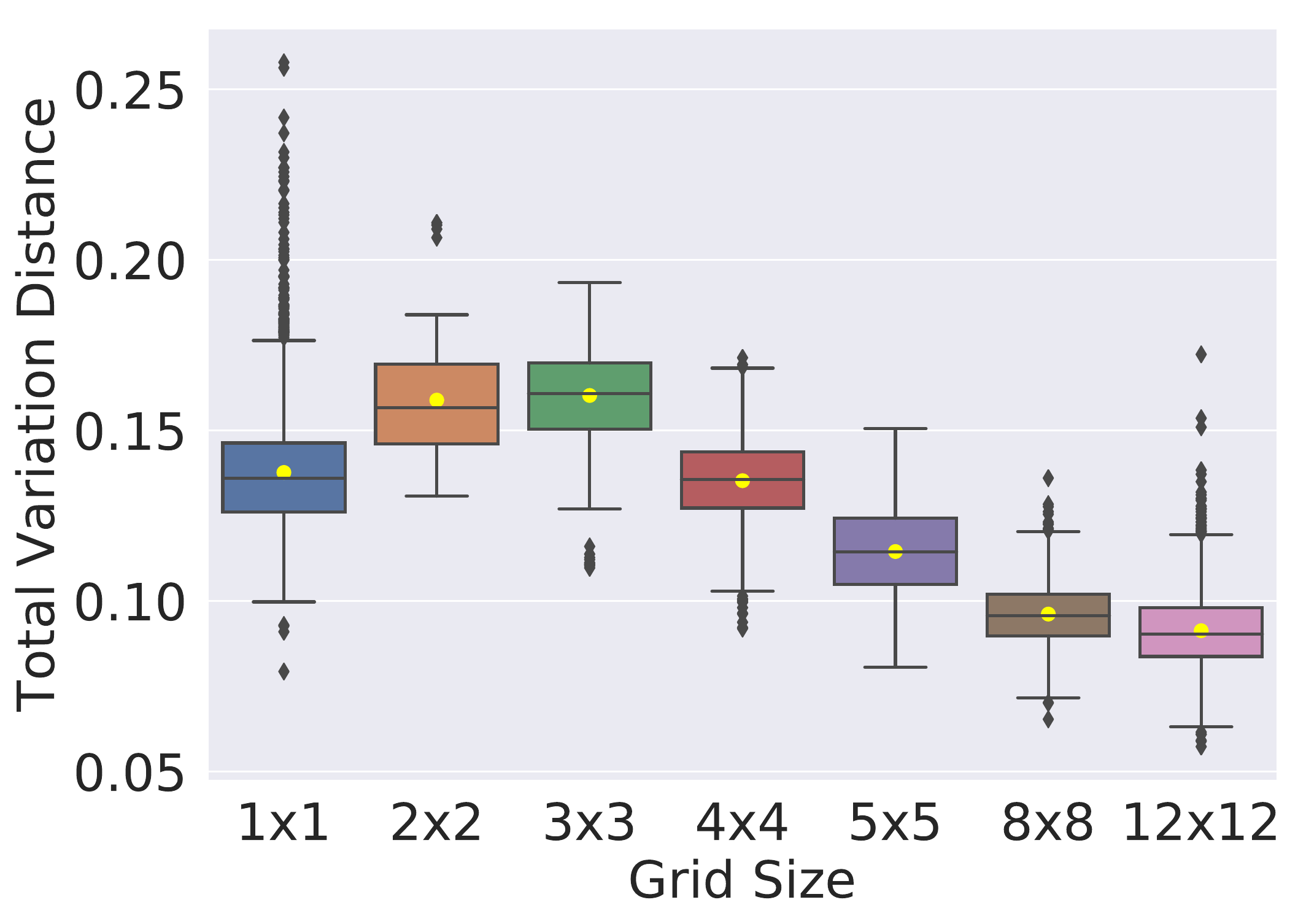}\label{fig:MNIST_tvd}}
	\subfloat[\scriptsize{\tiny Classes distribution for $1 \times 1$}]
	{\includegraphics[width=0.249\textwidth]{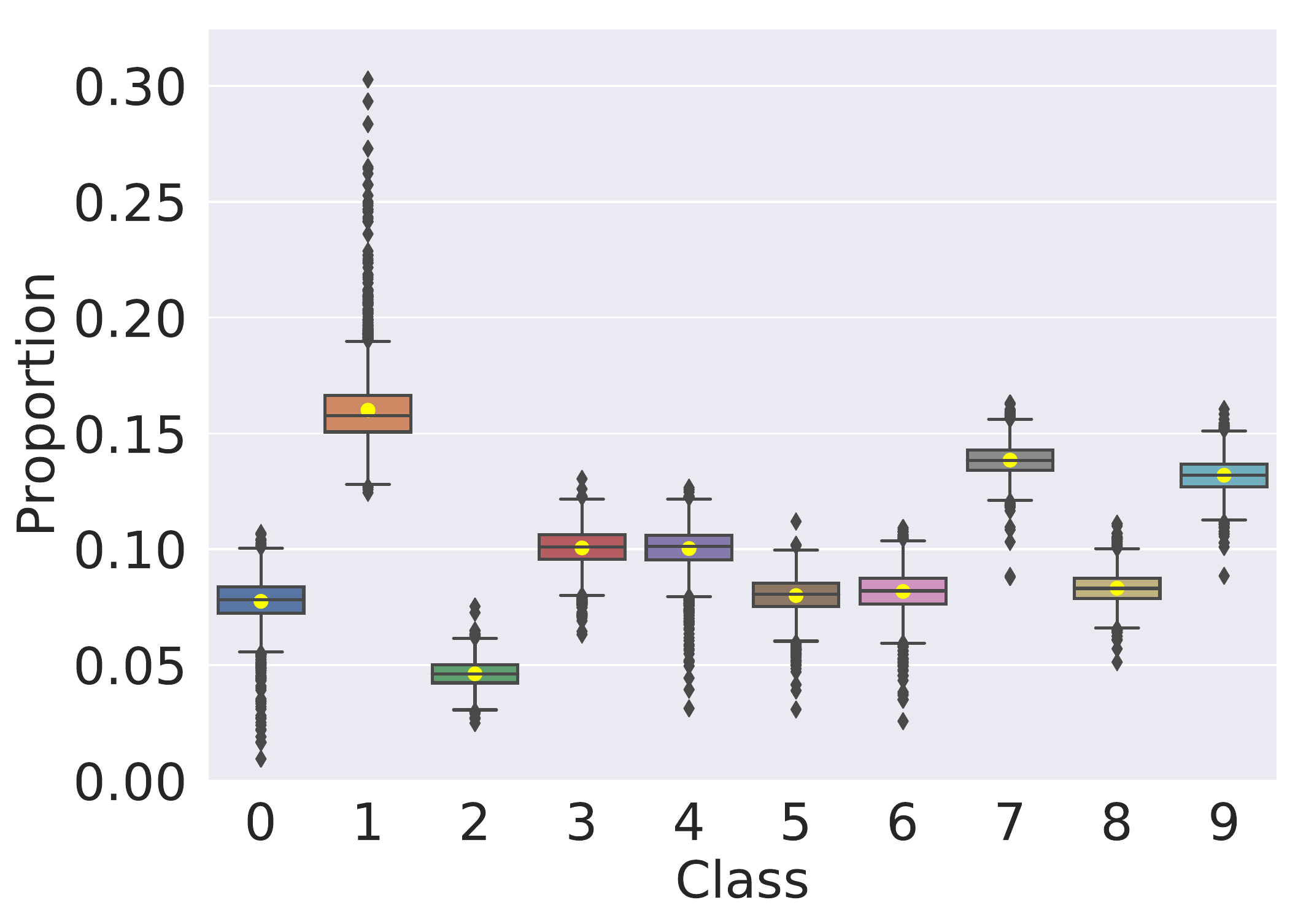}\label{fig:MNIST_distribution_1x1}}
	\subfloat[\scriptsize{\tiny Classes distribution for $4 \times 4$}]
	{\includegraphics[width=0.249\textwidth]{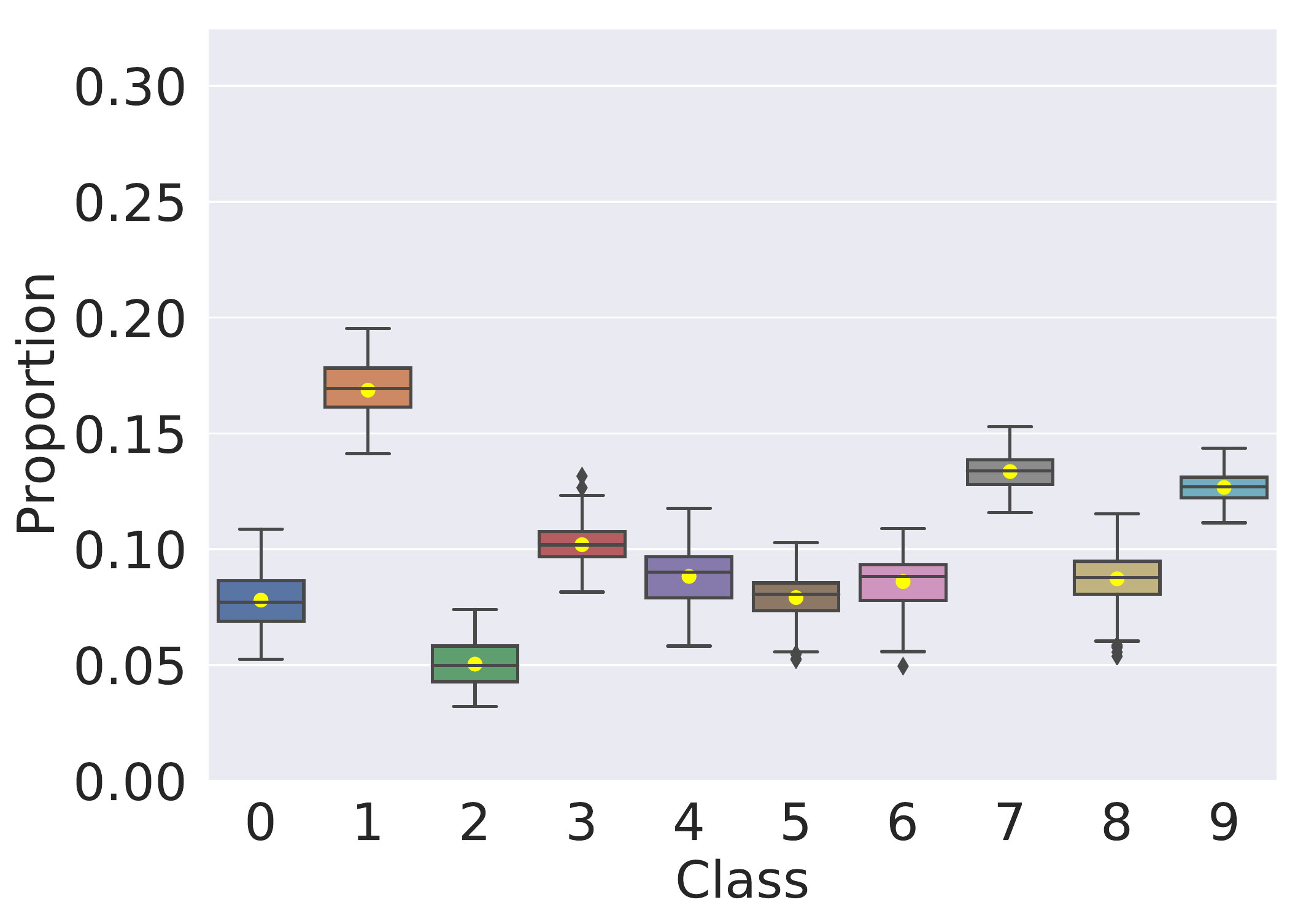}\label{fig:MNIST_distribution_4x4}}
	\subfloat[\scriptsize{\tiny Classes distribution for $12 \times 12$}]
	{\includegraphics[width=0.249\textwidth]{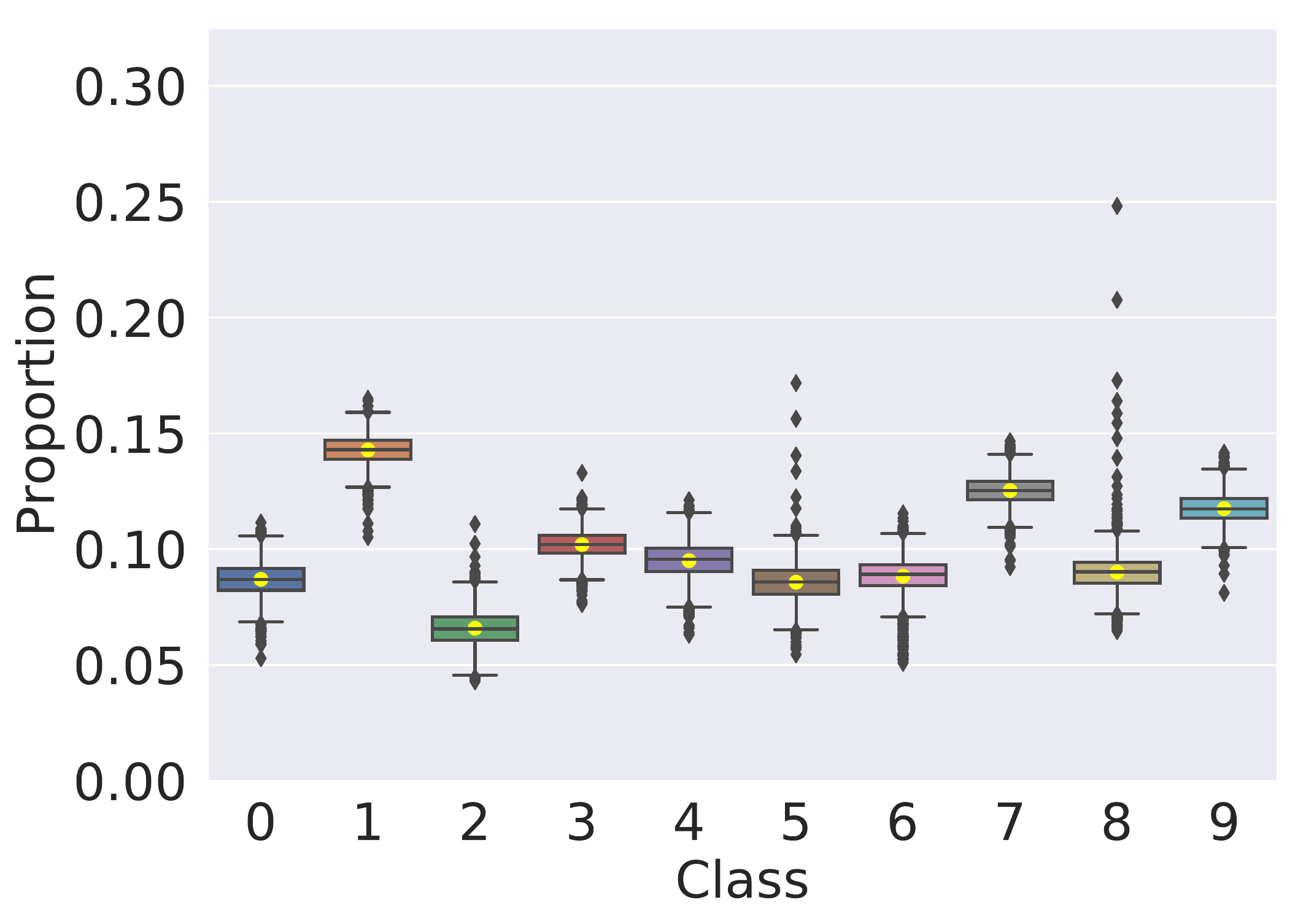}\label{fig:MNIST_distribution_12x12}}
	\captionof{figure}{\footnotesize Generator mixture distribution for MNIST. The average
          TVD is shown in Figure~\ref{fig:MNIST_tvd}. The larger grid sizes have lower TVD,
          which indicate that mixtures from larger grid sizes produce a more diverse set of images
          spanning across different classes. This is further supported by visualizing the distribution
          of each classes of generated images for different grid sizes. The distribution of each classes
          of generated images for $1 \times 1$ is in
          Figure~\ref{fig:MNIST_distribution_1x1}. The $4 \times 4$,
          Figure~\ref{fig:MNIST_distribution_4x4}, show a more uniform distribution. The
          distribution for $12 \times 12$, Figure~\ref{fig:MNIST_distribution_12x12}, is
          the most uniform.}
	\label{fig:MNIST_tvd_all}
	\end{minipage}\hspace{0.1em}
	
		\begin{minipage}[c]{\textwidth}
			\centering
			\includegraphics[width=0.35\textwidth]{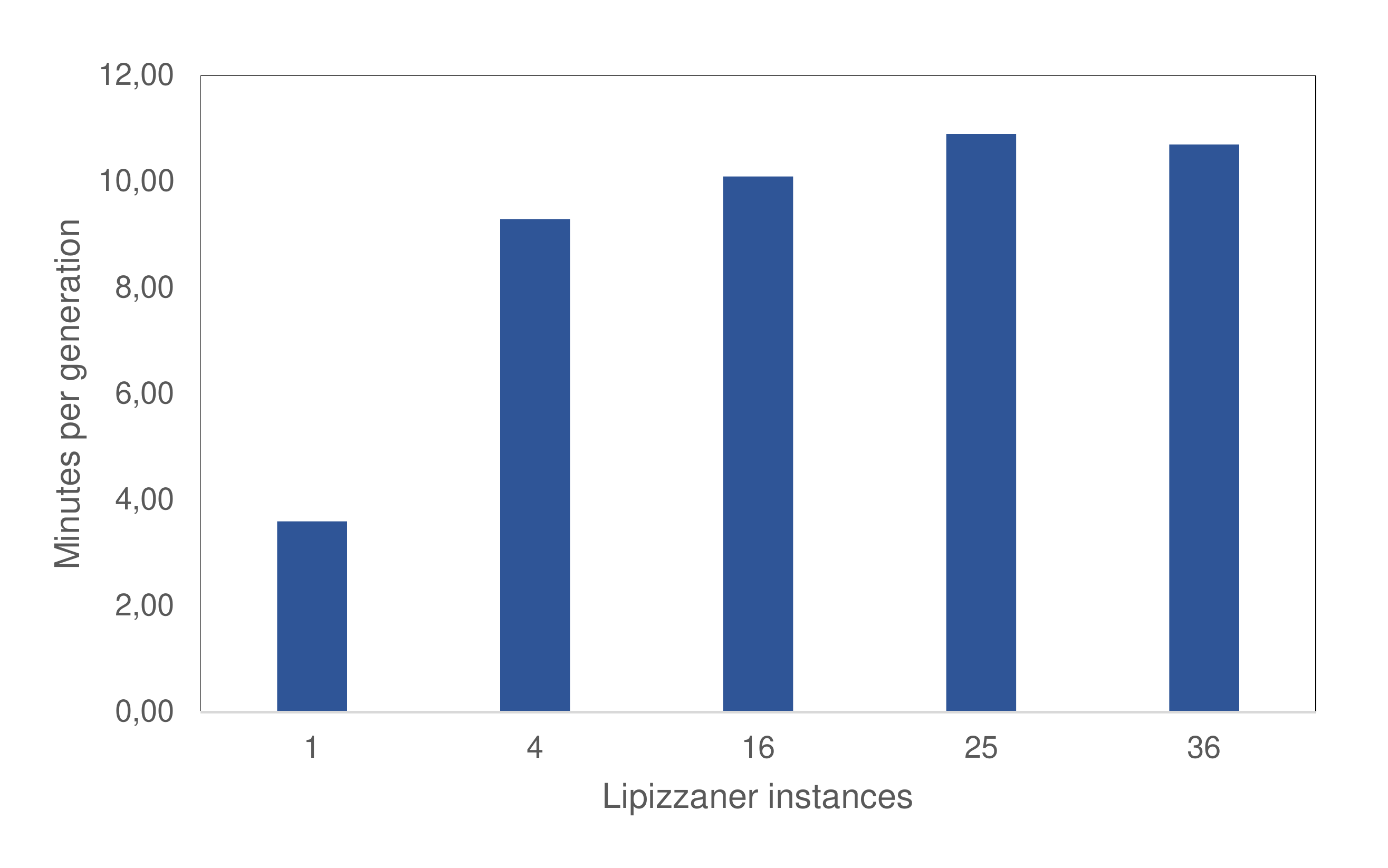}
			\captionof{figure}{\footnotesize Near constant training times on AWS per
				iteration on the CelebA dataset, averaged over 30
				iterations. X-axis shows the number of \horse instances and y-axis shows the
				duration in minutes per iteration.}
			\label{fig:scalability}
		\end{minipage}
\end{figure}

\textit{Generator Mixture Distribution.} We study the distribution of
the generator mixture. We follow \cite{Li2017Distributional} and
report the total variation distance (TVD) for the different grid
sizes, see Figure~\ref{fig:MNIST_tvd}. The larger grid sizes have
lower TVD, which indicate that mixtures from larger grid sizes produce
a more diverse set of images spanning across different classes. The
distribution of each classes of generated images for $1 \times 1$ is
in Figure~\ref{fig:MNIST_distribution_1x1} and is the least
uniform. For the $4 \times 4$,
Figure~\ref{fig:MNIST_distribution_4x4}, the distribution is more
uniform. Finally, the distribution for $12 \times 12$,
Figure~\ref{fig:MNIST_distribution_12x12}, is the most uniform.

\subsection{Celebrity Faces Dataset}

The CelebA dataset \cite{liu2015faceattributes} contains 200,000
portraits of celebrities and is commonly used in GAN
literature. \horse can overcome mode and discriminator collapses even
when only the smallest possible grid size ($2 \times 2$) is used. The
increased diversity is sufficient to replace collapsed individuals in
the next iteration, and even allows the system to prevent collapse in
most runs. An example for a recovering system is shown from iteration
25 to 28 in Figure~\ref{fig:dashboard}. The scaling performance for
this data set is different, and has different computational
requirements, so we are only able to measure the generative
performance up to a $4 \times 4$ grid. The results show no statistical
significant difference: \textbf{$1 \times 1$ (10 runs)} gives
$31.89\pm 1.26$, \textbf{$2 \times 2$ (10 runs)} gives $30.27\pm 0.50$
and \textbf{$4 \times 4$ (10 runs)} gives $30.59\pm 1.03$.

\textit{Scalability and Training Time.}
Scalability was one of the main requirements while designing
\horse. The spatial grid distribution architecture, allows the
computational effort to increase linearly instead of quadratically (up to $6 \times 6$ grid).
This claim is supported by the chart shown in
Figure~\ref{fig:scalability}, which illustrates a near linear training
time per iteration for different numbers of connected instances. The
initial relatively large step from one to four instances is caused by
the fact that multiple instances were run per GPU for the
distributed experiments; this increases the calculation effort per
GPU, and therefore affects the training time as well. We also observed low communication durations in our experiments:
exchanging data between two clients only takes 0.5 seconds on average
in state-of-the-art Gigabit Ethernet networks and is only performed
once per iteration. Additionally, the asynchronous communication
pattern leads to the usage of different time slots and therefore
reduces high network peak loads. The experiments were computed on AWS GPU cloud instances. Each instance
had one Nvidia Tesla K80 GPU with 12 GB RAM, 4 Intel Xeon cores
with 2.7 GHz each, and 60 GB RAM. The times shown are
averaged over 30 iterations of training a DCGAN neural network pair on
the CelebA dataset. The instances hosted Docker containers and
connected through a virtual overlay network.

%% file: conclusion.tex
\section{Conclusion}
\label{sec:conclusion}

\horse yields promising results in the conducted experiments and is
able to overcome otherwise critical scenarios like mode and
discriminator collapse. The main advantage of incorporating GANs in
coevolutionary algorithms is the usage of populations and 
therefore increased diversity among the possible solutions. Using a
relatively small spatial grid is sufficient to overcome the common
limitations of GANs, due to the spatial grid and asynchronous
evaluation. The performance also improves with increased grid size. In
addition, \horse scales well up to the grid sizes elaborated in the
conducted experiments (i.e. a grid size of $12 \times 12$ for MNIST and $6 \times 6$ for CelebA). Future work includes extending the GAN trainers used (e.g., WGAN),  investigating coevolutionary variants, and improving the
dashboard for tracing solutions over time.